\documentclass[a4paper, conference]{IEEEtran}      %

\IEEEoverridecommandlockouts                              %

\pdfoutput=1
\usepackage{amsmath}
\usepackage{bm}
\usepackage{booktabs}
\usepackage{caption}

\usepackage{graphicx}
\usepackage{subcaption}
\usepackage{epstopdf}

\usepackage{hhline}
\usepackage{leftidx}
\usepackage{makecell} %
\usepackage{mathtools} %
\usepackage{multirow}
\usepackage{textcomp}
\usepackage[pagebackref=true]{hyperref}
\usepackage[noadjust]{cite}

\title{\LARGE \bf
Segway DRIVE Benchmark: Place Recognition and SLAM Data Collected by A Fleet of Delivery Robots}

\author{Jianzhu Huai, Yusen Qin, Fumin Pang, Zichong Chen%
\thanks{This benchmark is supported by Segway Robotics Inc., Beijing, China.
        {Web site: www.segwayrobotics.com}. Authors can be contacted at \emph {firstname.lastname@ninebot.com}.}%
}

\begin{document}

\maketitle
\thispagestyle{empty}
\pagestyle{empty}

\begin{abstract}
	Visual place recognition and simultaneous localization and mapping (SLAM) have recently 
	begun to be used in real-world autonomous navigation tasks like food delivery. 
	Existing datasets for SLAM research are often not representative of in situ operations,	leaving a gap between academic research and real-world deployment.
	In response, this paper presents the Segway DRIVE benchmark, a novel and challenging dataset suite collected by a fleet of Segway delivery robots. 	
	Each robot is equipped with a global-shutter fisheye camera, 
	a consumer-grade IMU synced to the camera on chip, two low-cost wheel encoders, 
	and a removable high-precision lidar for generating reference solutions.
	As they routinely carry out tasks in office buildings and shopping malls while collecting data, 
	the dataset spanning a year is characterized by planar motions, moving pedestrians in scenes, 
	and changing environment and lighting. 
    Such factors typically pose severe challenges and may lead to failures for SLAM algorithms.
	Moreover, several metrics are proposed to evaluate metric place recognition algorithms. 
	With these metrics, sample SLAM and metric place recognition methods were evaluated on this benchmark.
	
	The first release of our benchmark has hundreds of sequences, covering more than 50 km of indoor floors. 
	More data will be added as the robot fleet continues to operate in real life. 
    The benchmark is available at \url{http://drive.segwayrobotics.com/#/dataset/download}.

\end{abstract}

\section{INTRODUCTION}
The demand for mobile robots with the autonomous navigation capability has intensified in recent years.
The online shopping and on-demand food delivery market in China has been 
growing at a rate of 30\%\texttildelow50\% per year,
leading to labor shortage and rising delivery cost.
Delivery robots have the potential to solve the dilemma caused by the growing consumer demand and decreasing delivery workforce.
Autonomous operation of such robots relies on place recognition and SLAM techniques 
that can successfully handle a variety of real-world scenarios.
By providing diverse testing scenarios, open benchmark datasets are crucial for researchers to identify challenging problems 
and to enhance place recognition and SLAM development.

\begin{figure}[thpb]
	\centering
	\subcaptionbox*{}{\includegraphics[width=0.32\columnwidth,trim={0 0mm 0 0mm},clip]{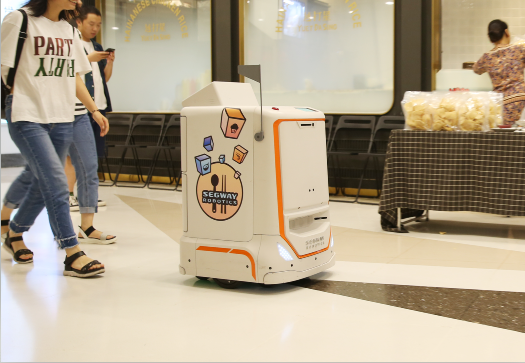}}
	\hfil
	\subcaptionbox*{}{\includegraphics[width=0.32\columnwidth,trim={0 0mm 0 0},clip]{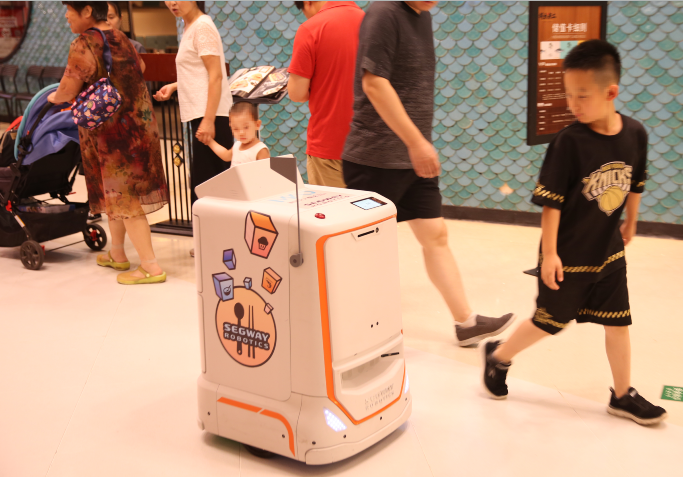}}
	\hfil
	\subcaptionbox*{}{\includegraphics[width=0.32\columnwidth,trim={0 0mm 0 13mm},clip]{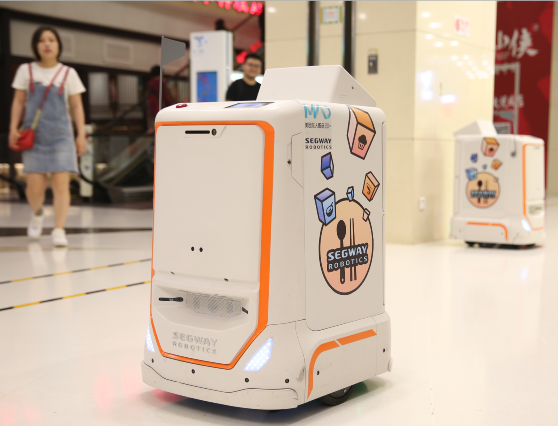}}
	\\[-2.5ex]
	\subcaptionbox*{}{\includegraphics[width=0.32\columnwidth,trim={0 0mm 0 0},clip]{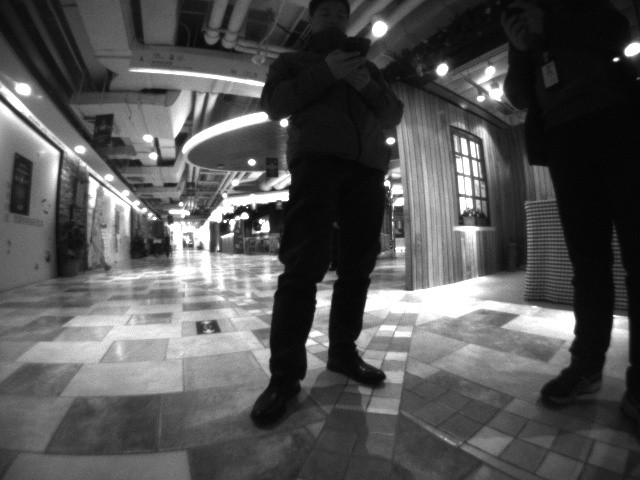}}
	\hfil
	\subcaptionbox*{}{\includegraphics[width=0.32\columnwidth,trim={0 0mm 0 0mm},clip]{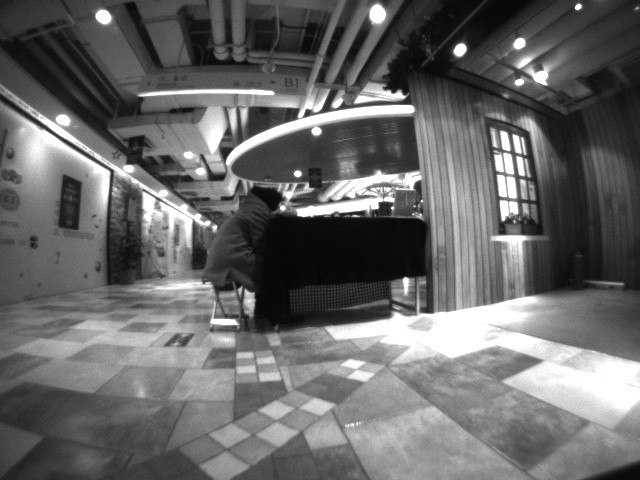}}
	\hfil
	\subcaptionbox*{}{\includegraphics[width=0.32\columnwidth,trim={0 0mm 0 0},clip]{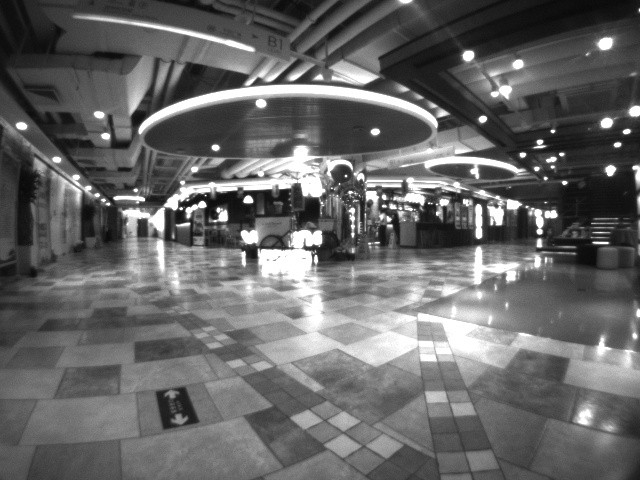}}
	\\[-2.5ex]
	\subcaptionbox*{}{\includegraphics[width=0.32\columnwidth,trim={0 0mm 0 0},clip]{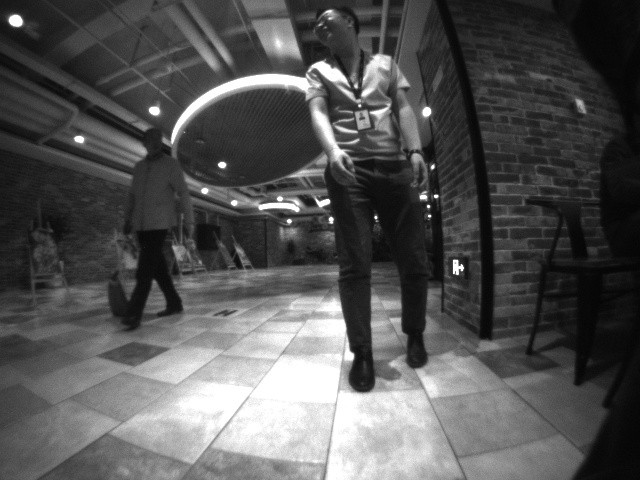}}
	\hfil
	\subcaptionbox*{}{\includegraphics[width=0.32\columnwidth,trim={0 0mm 0 0mm},clip]{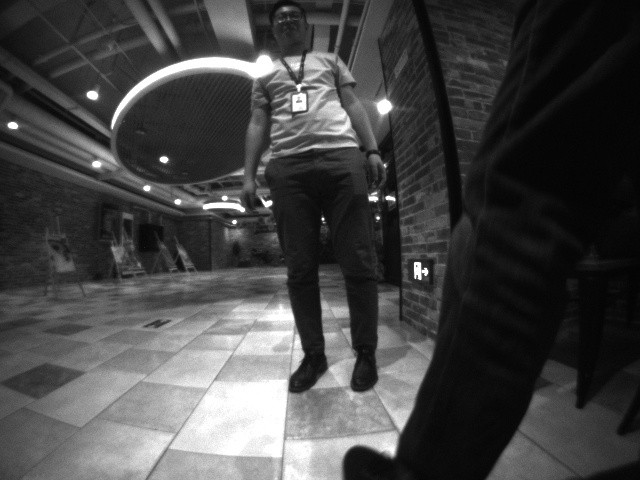}}
	\hfil
	\subcaptionbox*{}{\includegraphics[width=0.32\columnwidth,trim={0 0mm 0 0},clip]{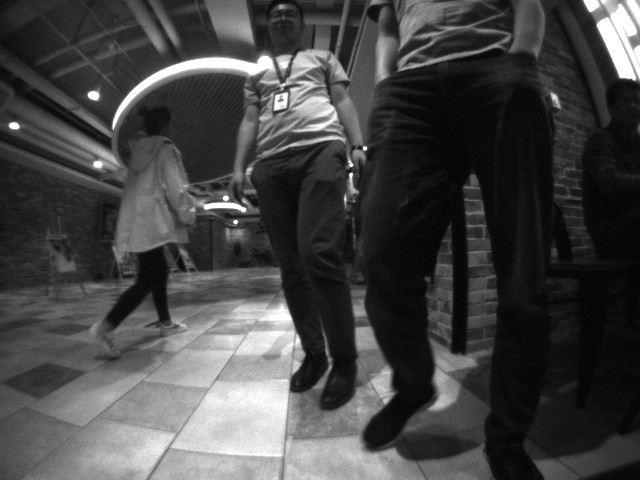}}
	\\[-2.5ex]
	\subcaptionbox*{}{\includegraphics[width=0.32\columnwidth,trim={0 0mm 0 0},clip]{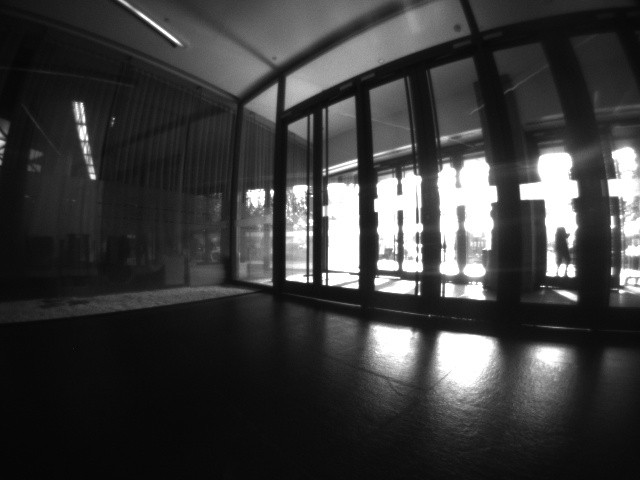}}
	\hfil
	\subcaptionbox*{}{\includegraphics[width=0.32\columnwidth,trim={0 0mm 0 0mm},clip]{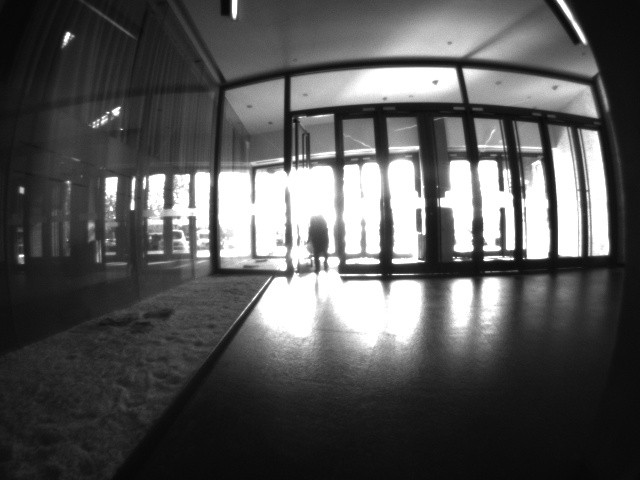}}
	\hfil
	\subcaptionbox*{}{\includegraphics[width=0.32\columnwidth,trim={0 0mm 0 0},clip]{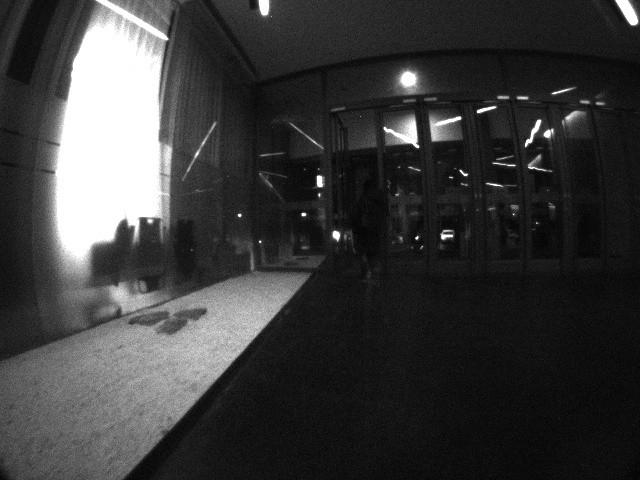}}
	
	\caption{First row shows a Segway delivery robot doing tasks in a highly dynamic shopping mall;
	Second row shows long-term structural changes;
	Third row shows people walking in front of the robot;
	Fourth row shows lighting changes at different times of day.}
	\label{fig:samples}
\end{figure}

In this paper, we present the Segway DRIVE Benchmark, a novel and challenging benchmark for place recognition and SLAM research.
The data were collected by a fleet of delivery robots while they were carrying out tasks,
such as delivering parcels and meals across office buildings and shopping malls (Fig.~\ref{fig:samples}).
In view of existing relevant benchmarks \cite{burri25012016}\cite{pfrommer2017penncosyvio}, 
this one presents many unique challenges faced by real-world robot deployment on a large scale:

\begin{itemize}
	\item Commodity inertial measurement units (IMUs) were used, meaning that their characteristics were less favorable for odometry algorithms than the industrial ones.
	
	\item Prior to deployment in a location, at least one data sequence for mapping was recorded to create a SLAM map for online place recognition.
    With the map, robots repeatedly performed the assigned tasks on similar routes in the location, recording test sequences.
    The mapping and testing procedures have been repeated for each location up to one year.
    These data were subject to lighting changes, e.g., at different times of a day,
    and environmental changes, e.g., by refurbishment,
    causing difficulties for visual place recognition and map fusion methods \cite{muhlfellner2016summary}.
    
	\item The data were recorded indoors seeing many moving people and objects which might hamper the feature tracking module typically found in SLAM techniques.
	
	\item Since our robots primarily moved on planar floors, 
	the motion pattern introduced additional unobservable states to a vision-inertial navigation system (VINS) 
	whose persistent unobservables were the global position and yaw angle \cite{wu2017vins}.
	To tackle the observability problem of the VINS for wheeled vehicles,
	the wheel odometry data were included in the benchmark.
\end{itemize}

The first release of our benchmark has about 100 sequences collected by robots repeatedly exploring 5 different indoor locations over a period of one year.
The sensors mounted on a robot for data capture include a RealSense visual inertial (VI) sensor, two wheel encoders, and a Hokuyo 2D lidar.
The ground truth poses (positions and orientations) generated from the lidar data have been provided in a single frame of reference for the majority of data sequences.
The benchmark will continue to expand as the robot delivery service proliferates.
Useful tools to interact with the data and evaluation scripts have been 
provided along with the data at \url{http://drive.segwayrobotics.com}.

On our benchmark, we evaluated several SLAM methods \cite{leutenegger2015keyframe}\cite{qin2018vins}\cite{bloesch2015robust} with metrics proposed in \cite{sturm12iros} using the ground truth poses.
To describe the performance of visual localization, we propose novel metrics that
assess both the frequency and outliers of relocalizations for a test sequence relative to a mapping sequence. These metrics were demonstrated with a state-of-the-art metric place recognition method \cite{muhlfellner2016summary} on the benchmark.

\section{RELATED WORK} 

A number of benchmarks have been released for evaluating place recognition or SLAM methods.
A few relevant ones are listed below in chronological order.

\textbf{KITTI visual odometry} \cite{geiger2012cvpr}:
An outdoor dataset suite with 22 stereo image sequences from a driving car 
with half of them having GPS/INS fused poses as ground truth.

\textbf{Malaga Urban} \cite{blanco2014malaga}:
An outdoor dataset suite with 15 stereo image sequences from a driving car without ground truth. The sequences did not cover the same location many times.

\textbf{EUROC MAV} \cite{burri25012016}:
An indoor dataset of 11 stereo visual and inertial sequences from a Micro Aerial Vehicle (MAV)
in indoor and outdoor locations with ground truth poses from a laser tracker or a motion capture system.

\textbf{NCLT} \cite{carlevaris2016university}: 
An indoor and outdoor dataset suite of 27 sequences collected by camera, lidar, GPS/IMU sensors on a Segway robot which repeatedly explored the campus over 15 months. Ground truth poses in a single reference frame by fusing GPS and lidar data were provided for all sessions.

\textbf{Oxford RobotCar dataset} \cite{maddern2017}:
An outdoor dataset consisting of images, lidar scans, and GPS/IMU data, collected
under all weather conditions as the RobotCar platforms traversed the same route many times in one year. 
The GPS/IMU fused solutions was provided but had varying accuracy.

\textbf{TUM VI} \cite{schubert2018tum}:
An indoor and outdoor dataset suite of 28 visual inertial sequences collected in controlled nearly static environments by hand-held stereo cameras and IMU sensors.
For all sequences, ground truth poses from a motion capture system were provided covering the beginning and the end.

\section{SENSOR SETUP}

\begin{figure}[thpb]
	\centering
	\includegraphics[scale=0.2,trim={2in 2in 2in 2in},clip]{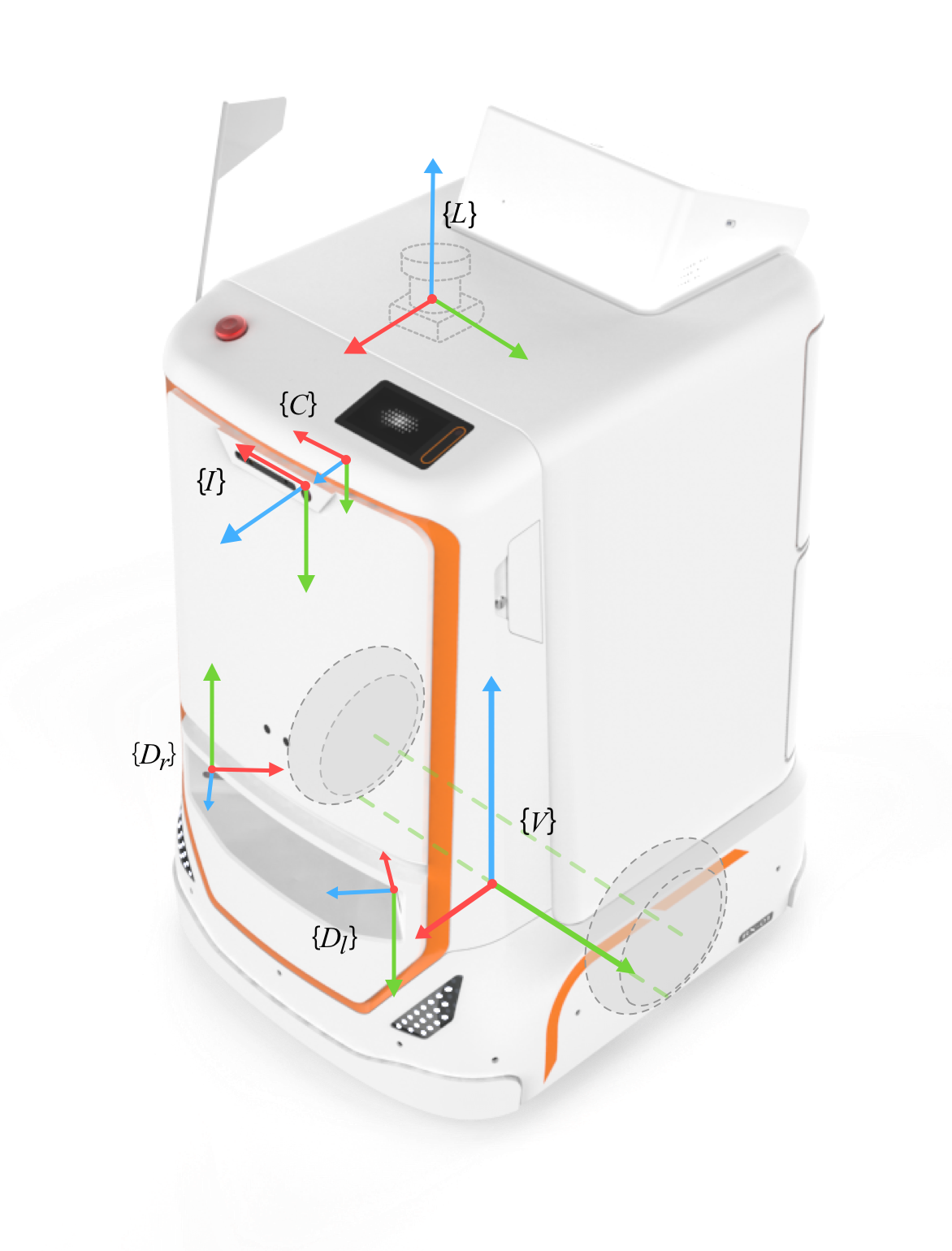}
	\caption{A Segway delivery robot drawn with the sensor coordinate frames:
		 top to bottom, \{$L$\} - lidar, \{$C$\} - camera, \{$I$\} - IMU,
		  \{$V$\} - vehicle}
	\label{robotcad}
\end{figure}

A Segway delivery robot, shown in Fig.~\ref{robotcad}, is outfitted with many consumer-grade sensors
for autonomous navigation and data collection.

\subsection{Hardware setup} 

The VI sensor (Intel RealSence ZR300) is mounted on the front panel of the robot.
It has a monochrome fisheye camera of a global shutter and a $166.5^{\circ}$ field of view (FOV).
The camera streams images at 30 fps, which are saved at 10 fps.
The IMU in the VI sensor, BMI055, logs accelerometer data at 250Hz and gyroscope data at 200Hz.
The IMU and the camera are synchronized in hardware, and the images are timestamped by the sensor at the middle of exposure.
For the visual and inertial data, the timestamps by both the VI sensor and the host computer are recorded.

Two encoders are installed on the wheels.
The data of wheel encoders are stamped with times of arrival to the host computer.

For creating ground truth, a Hokoyo UTM-30LX single beam lidar with $270^{\circ}$ FOV, 30 m range, and $\pm$ 50 mm accuracy, is temporarily mounted on the robot to collect laser scans logged by an external computer. Its synchronization to the host computer is discussed in Section ~\ref{ref_ground_truth}.

\subsection{Coordinate frame convention} 

In our dataset, several right-handed coordinate frames tied to sensors 
as shown in Fig.~\ref{robotcad} are defined, including \{$C$\} on the camera, \{$I$\} on the IMU, \{$V$\} on the vehicle, and \{$L$\} on the lidar. 

The \{$C$\} frame is tied to the fisheye camera, the \{$I$\} frame to the accelerometer triad,
of the VI sensor module following the ROS convention as in \cite{huai2017collaborative}.
The vehicle frame, \{$V$\}, has x pointing forward along the chassis,
 y left along the wheel axis, z up along the trunk.
Its origin is the point where the wheel axis center projects along the z-axis 
onto the ground.
\{$L$\} is attached to the center of the lidar with x pointing forward along the 
bisector of a scan's field of view, y to the left of the sensor, and z up.
For calculating covariance of planar coordinates, a pseudo-camera frame \{$C_p$\} is introduced. 
Its origin is at \{$C$\} and its orientation is the same as \{$V$\},
thus its z-axis is opposite to the gravity direction in general.
As sensors move about over time, their associated frames may be timestamped for clarity.
For instance, \{$I(t_j)$\} denotes an Earth-fixed IMU frame at epoch $t_j$.

The 6 degree-of-freedom (DOF) pose of frame \{$X$\} expressed in frame \{$Y$\} is defined as a SE(3) element, 
$T_{YX}=\left [ \begin{matrix}
R_{YX} & \bm t_{YX}\\ 
\bm 0 & 1
\end{matrix} \right ]$. The transformation $T_{YX}$ transforms point coordinates $ \leftidx{^X}{\bm p} \in {R}^3 $ expressed in \{$X$\} to those in \{$Y$\}, $ \leftidx{^Y}{\bm p}$, through $ \left[\begin{matrix}
\leftidx{^Y}{\bm p} \\
1
\end{matrix}\right] = T_{YX}\left[\begin{matrix}
\leftidx{^X}{\bm p} \\
1
\end{matrix}\right]$. For internal computations, rotations like $R_{YX}$ are
 also expressed by Hamilton quaternions as in \cite{huai2017collaborative}.
As the robot travels on planes in general and world frames and body frames are usually defined with z-axis pointing along the negative gravity, 
a transform between a world frame \{$W$\} and a body frame \{$B$\} is often expressed by longitudinal (x) and transverse (y) translations and 
the yaw angle ($\theta$), i.e.
$T_{WB}=\left [ \begin{matrix}
R_z(\theta) & [x, y, 0]^T\\ 
\bm 0 & 1
\end{matrix} \right ]$. Conveniently, we define $\pi(\cdot)$ such that $[x, y, \theta]^T = \pi(T_{WB})$.

\subsection{Sensor calibration}\label{ref_sec_calib}

The intrinsic parameters of the fisheye camera and its extrinsic parameters relative to the IMU obtained by the Kalibr toolbox \cite{furgale2013unified} are provided with each dataset.

The factory-calibrated deterministic error parameters of the IMU are stored in the device and used to correct the measured values before transmission.
The stochastic errors of the 3 accelerometers and 3 gyroscopes of the IMU were
characterized in our workplace by the Allan variance analysis.
Since the computed noise characteristics for up to 5 IMUs in our experiments are close to their nominal values, 
their averages are shared by the data collected by robots.
Sample analysis results and recommended noise characteristic values for several SLAM algorithms \cite{leutenegger2015keyframe}\cite{qin2018vins}\cite{bloesch2015robust}
are presented on the benchmark website.

For the two encoders on the wheels which realize the \{$V$\} frame, 
the extrinsic parameters between \{$V$\} and \{$C$\} are obtained from the CAD 
drawing of the robot, i.e., $T_{VC}$.
These extrinsic parameters are used to convert the encoder measurements 
to robot poses.
In particular, tick measurements from the two wheel encoders are integrated 
to 2D odometry poses at the timestamps of the images following a differential 
steering model \cite{nikolauscorrell2016}.
Then, the 2D poses expressed in \{$V$\} are converted to 3D poses, 
$T_{V(t_0) V(t_j)}$, assuming a zero translation in the z-axis and zero 
rotations about the x- and y-axis, 
and finally transformed to the \{$C$\} 
frame with $T_{VC}$. Here $t_0$ denotes the start epoch of a dataset capture session.
The resulting poses, $T_{C(t_0) C(t_j)}$, are provided in each dataset.

The time offset of the Hokuyo lidar to the host computer and its extrinsic parameters are estimated as detailed in Section ~\ref{ref_ground_truth}.

\section{DATASET}

This section overviews the content, format, and ground truth of our benchmark.

\subsection{Sequences} 

Out of thousands of data sequences in the Segway Robotics database, 
nearly 100 sequences for 8 locations, collected in a period of 6 months, 
have been made available as summarized in Table ~\ref{seqtable}.

\begin{table}
\caption{Dataset sequence overview}
\begin{tabular}{lccc}
\hline
\textbf{Location} & \textbf{\#Sessions}  & \textbf{\begin{tabular}{@{}c@{}}\#Sessions \\ with 2D lidar\end{tabular}} & \textbf{Earliest \texttildelow\enskip Latest} \\
\hline
B2\_F1 & 18 & 17 & 2018-09-21 \texttildelow\enskip 2018-12-27 \\
B6\_B1 & 29 & 26 & 2018-08-02 \texttildelow\enskip2018-12-21\\
B6\_F1 & 5 & 5 & 2018-11-13 \texttildelow\enskip2018-12-10 \\
B6\_F5 & 7 & 7 & 2018-08-21 \texttildelow\enskip2018-12-14\\
joycity\_F7 & 2 & 2 & 2018-09-12 \texttildelow\enskip2018-09-12\\
joycity\_F8 & 2 & 1 & 2018-09-12 \texttildelow\enskip2018-09-12\\
lvdi\_B1 & 7 & 4 & 2018-09-06 \texttildelow\enskip2018-10-11 \\
lvdi\_F1 & 4 & 2 & 2018-09-13 \texttildelow\enskip2018-10-11 \\
\hline
\label{seqtable}
\end{tabular}
\end{table}

These data sequences highlight many visual localization and mapping challenges listed below.

\textbf{Stationary periods}: The robot stays stationary for a period in which a SLAM method is unable to estimate the scene depth with the given sensors.

\textbf{Rapid rotation}: The robot makes sharp turns abruptly which may lead to motion blur in images and wheel slips.

\textbf{Dynamic environment}: Moving objects and pedestrians may adversely affect localization precision.

\textbf{Repetitive scenes}: The office buildings and shopping malls often have similar structures and decorations in different places, which may lead to wrong place recognition.

\textbf{Environment and illumination variation}: For one location, many data sequences are recorded at different times of day and in a period of half an year. The involved environmental and lighting changes are challenging to relocalization methods.

\textbf{Rough terrain}: Robots running on bumpy floors may observe choppy IMU signals.

\textbf{Less textured scenes}: The robot may see a textureless wall or an expansive hall, causing a visual odometry method to drift.

\textbf{Reflections and shadows}: Shiny floors, glasses, and mirror-like objects may cause wrong feature associations in a visual odometry method.

\subsection{Data Format}

For each sequence one ROS bag file is provided with topics shown in Table~\ref{formattab}.
As a robot moves with a speed about 1 m/s, fisheye camera images are typically captured at 10Hz
which is deemed sufficient for localization in our indoor delivery applications.
The raw accelerometer data are logged at 250Hz, and the raw gyro data at 200Hz from the VI sensor.
The raw accelerometer data are interpolated online at epochs of gyro data, resulting in IMU data at 200 Hz.
The data captured by the two wheel encoders are used to derive the fisheye camera poses in an Earth-fixed frame, 
$T_{C(t_0) C(t_j)}$, as discussed in Section ~\ref{ref_sec_calib}, 
serving as the odometer input.

\begin{table}
\renewcommand\arraystretch{1.5}
\caption{List of topics in a dataset ROS bag}
\begin{tabular}{ll}
\hline
\thead{\textbf{Topic}}  & \thead{\textbf{Description}}\\
\hline
/cam0/image\_raw & Fisheye camera data at \texttildelow 10Hz\\
\hline
/imu0 & \makecell[cl]{IMU data at 200Hz, expressed in \{$I$\}}\\
\hline
/tf0 & \makecell[cl]{Camera poses at \texttildelow 10Hz derived from the wheel \\ encoder data, expressed in an Earth-fixed frame}\\
\hline
\label{formattab}
\end{tabular}
\end{table}

Along with the ROS bag, the calibration information is provided in a \texttt{yaml} file, 
including intrinsic and extrinsic parameters of the VI sensor 
and IMU noise characteristics.

The ground truth for a data session, if available, is provided in a \texttt{csv} file.
Essentially, it is the fisheye camera poses in an Earth-fixed world frame 
derived from the lidar data as discussed next.

\subsection{Ground truth}\label{ref_ground_truth}

To evaluate the performance of a tracking or mapping algorithm, 
it is common to measure the agreement of its output sequential poses with
 a reference trajectory \cite{sturm12iros}\cite{geiger2012cvpr}. 
For our data captured with nearly planar motion, the 2D ground truth trajectories are
generated from laser scans captured by a Hokuyo lidar mounted rigidly relative to the VI sensor.
The laser scans are processed offline by the approach proposed in \cite{hess2016real} with loop closure at a 5cm resolution.
For pairs of distinct points in the resultant laser map, we spot-checked distances estimated from the lidar data against the values obtained with a tape measure. 
Even for distances up to 10 m, the deviations from the tape measure were less than 10 cm.

Since the lidar is connected to another computer than the one for the other sensors, the lidar data need to be aligned in time to the VI data.
This is done by correlating norms of angular rates sampled from the SLERP curves fitted to the lidar trajectory and those of gyroscope angular rate samples \cite{furrer2017FSR}.
Given the time offset, the extrinsic parameters of the lidar relative to the fisheye camera is estimated by the method from \cite{guo2012analytical} implemented in the camodocal toolbox \cite{heng2013camodocal}.
Finally, both the time offset and extrinsic parameters of the lidar are refined at once with the oomact toolbox \cite{sommer2015continuous}.
With these parameters, the lidar reference solution, $T_{W_L L}$, is shifted in time and transformed to the fisheye camera frame,
resulting in a sequence of $T_{W_L C}$ with \{${W_L}$\} being an Earth-fixed frame 
used in generating the lidar solution \cite{hess2016real}.
Consequently, in measuring the quality of poses estimated by a SLAM algorithm, 
the evaluation procedure can be carried out with minimal user intervention using the provided tools in python.

\section{EVALUATION}

A plethora of visual SLAM and place recognition algorithms have been developed over recent years.
Many of them fall into three categories, 
odometry algorithms which estimate incremental motion, 
mapping algorithms which create an optimized map of an environment, 
and place recognition algorithms which recognize previously observed scenes 
and optionally provide metric location estimates \cite{hloc2018}.
To show that our dataset serves well to benchmark odometry, mapping, and metric place recognition algorithms under planar motion,
this section describes evaluation metrics and showcases their use with test algorithms on the dataset.

\subsection{Error metrics}

We use two metrics proposed in \cite{sturm12iros} to evaluate the performance of an odometry or mapping algorithm.
Both metrics are calculated from discrepancies between the estimated poses and the ground truth trajectory. 
The first, relative pose error (RPE), examines the local pose error over a fixed time window, 
and thus is useful for evaluating odometry methods. 
Just as a pose has the translation and the rotation part, the RPE metric is computed 
in terms of relative translation error (RTE) and relative rotation error (RRE).
The second, absolute trajectory error (ATE), looks into the absolute distances between the estimated and the reference trajectory, 
and thus is suitable to evaluate the global consistency of poses estimated by mapping methods.

For metric place recognition methods, we propose to evaluate them in terms of the number of valid localizations and 
the variance of localization occurrences over fixed distance intervals using the wheel odometry.
A place recognition algorithm often begins by learning an area from the mapping data
which is used to create a metric summary map \cite{muhlfellner2016summary} or 
to form a database of images with computed coordinates \cite{kendall2015convolutional}.
Against the learned data, metric localizations can be obtained by
querying images sampled from the test data with the place recognition method,
 accumulating into number of positive localizations, $N_p$. 

A false positive is determined by comparing the relative motion between two consecutive localizations 
and its counterpart from the wheel odometry
inspired by the reference relation concept in \cite{kummerle2009measuring}.
The validity of this test depends on two assumptions. 
First, false localizations are assumed to occur independently. 
This means, if two consecutive localizations are both false, the relative motion between them is unlikely to agree with that obtained from the wheel odometry. 
This assumption may break if a sequence of query images is used in determining one localization.
Secondly, the differential drive model of the wheels used to propagate the pose uncertainty is assumed to apply well to our robots.
This may become invalid if a wheel slips too much.
Alternatively, the lidar solution, \{$T_{W_L C}$\}, can be used in place of the wheel odometry, \{$T_{C(t_0) C}$\}, 
as a reference trajectory for test data with laser scans.

The idea is formulated as follows. Let's denote a reference pose by $T_{WC}$ where \{$W$\} is an Earth-fixed frame, e.g., \{$W_L$\}. 
For two consecutive localizations expressed in a global frame \{$G$\} used in place recognition,
$T_{G C(t_j)}$ and $T_{G C(t_{j+1})}$, the reference odometry poses at their timestamps, 
$T_{W C(t_j)}$ and $T_{W C(t_{j+1})}$, can be found by interpolation in SE(3).
The difference between the two relative motions and its covariance are computed as below.
\begin{multline}
\Delta T \equiv [\Delta x, \Delta y, \Delta\theta]^T \\
= \pi([T_{G C(t_j)}T_{CC_p}]^{-1}T_{G C(t_{j+1})}T_{CC_p}) \\
- \pi([T_{W C(t_j)}T_{CC_p}]^{-1}T_{W C(t_{j+1})}T_{CC_p})
\end{multline}
\begin{multline}
cov(\Delta T) = 2 cov(\pi(T_{G C(t_j)}T_{CC_p}) \\
+ cov(\pi(T_{C_p(t_j)C_p(t_{j+1})}))
\end{multline}

In the first term $cov(\pi(T_{G C(t_j)}T_{CC_p}))$ is the covariance of a localization, 
empirically set to $diag(0.1^2, 0.1^2, (5\pi/180)^2)$. 
The second term, covariance of the relative odometry, is computed by uncertainty propagation with the differential-drive process model (pp 50-54 of \cite{nikolauscorrell2016}).
For short distances, the first term typically dominates the second term,
 overshadowing the effect of wheel slips.

The number of false positives $N_{fp}$ is estimated as half of the count of $\Delta T$'s that exceed the Mahalanobis threshold.
\begin{equation}
N_{fp} = \frac{1}{2} \left\vert\{\Delta T \:\vert\: \sqrt{\Delta T^T cov(\Delta T)^{-1} \Delta T} > \lambda_{3, 0.025} \}\right\vert
\end{equation}
where $\lambda_{3, 0.025}$ is the upper critical value for $\chi^2$ distribution with 3 degrees of freedom at significance level 0.025 \cite{nist2012}.

Thus, the number of valid localizations (true positives) is $N_{tp} = N_{p} - N_{fp}$. 
For better comparison, this number can be normalized by the traveled distance or the time span.
\begin{align}
&\begin{aligned}
\mathllap{\dot{N}_{tp}} & = N_{tp} / t_n
\end{aligned}\\
&\begin{aligned}
\mathllap{\bar{N}_{tp}} & = N_{tp} / L
\end{aligned}
\end{align}
where $n$ is the last image's index in the test data, 
$L$ is the total traveled distance.
For a given pair of mapping and test sessions, the two metrics indicate
the ability of a place recognition module to correctly associate present and past observations.

Another metric, the standard deviation of place recognition frequencies over fixed spatial distance intervals,
quantifies the regularity of localizations over the course of a test data. 
For a test sequence covering a distance of $L$, 
the standard deviation $s_{PRF}$ is computed over the localization frequencies for contiguous intervals of length $\Delta$.

\begin{equation}
s_{PRF} = \frac{1}{N_{tp}/m}\sqrt{\frac{1}{m}\sum_{i=0}^{m-1}\left(l_i-\frac{N_{tp}}{m}\right)^2}
\label{varloc}
\end{equation}
Here $l_i$ denotes the number of true localizations in the interval $\left[i\Delta, (i+1)\Delta\right)$ and $m=\left \lceil L/\Delta \right \rceil$.

\subsection{Benchmark odometry algorithms}

Using ATE and RPE, this section evaluates several state-of-the-art visual inertial odometry methods on some of our dataset,
in order to show that our dataset is useful for evaluating odometry methods and that incorporating the wheel encoder data improves the tracking accuracy.

\begin{figure*}[htb]
  \centering
  \includegraphics[width=1\linewidth,trim={0 0 0 2.55mm},clip]{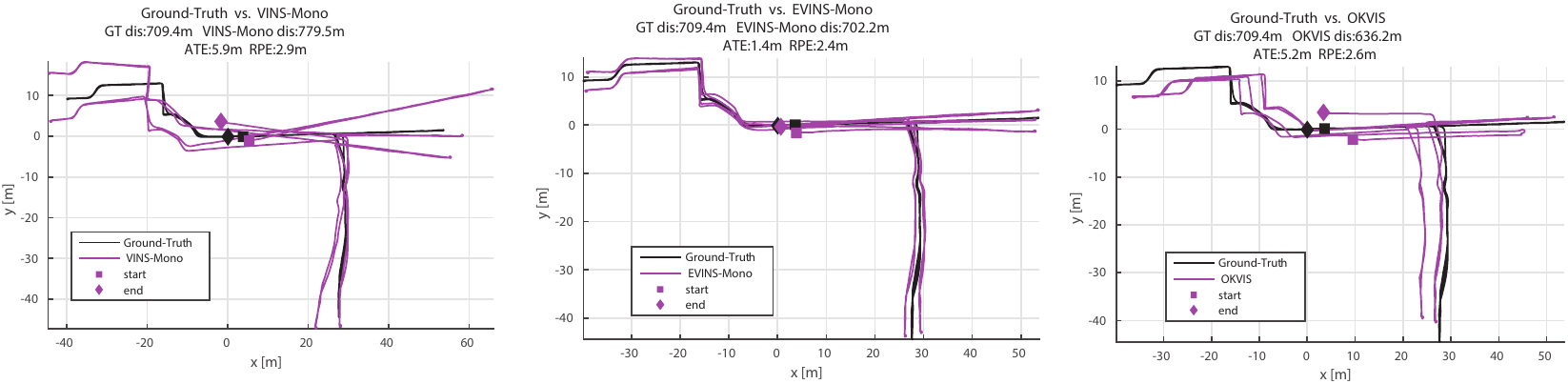}
  \caption{Visual odometry results on data 2018-08-02\_18-42-03 of B6\_B1.
  	 As the robot travels for 709.4 m, both the VINS-Mono (left) and OKVIS (right) algorithms drift much larger in terms of ATE and RPE than the extended VINS-Mono (middle) which uses wheel encoder data.}
  \label{viobenchmark}
\end{figure*}

The open source visual odometry programs, OKVIS \cite{leutenegger2015keyframe} and VINS-Mono \cite{qin2018vins} were chosen to run on our benchmark.
Moreover, we tested an extended VINS-Mono which could 
use the additional wheel encoder data as factors to 
constrain the relative motion between image frames in optimization.
We had also tried the ROVIO method \cite{bloesch2015robust} on our dataset, 
whose results were not presented as it often diverged halfway for a majority of our dataset.
In running VINS-Mono, its loop closure was disabled for better comparison.
For a specific odometry method, the intrinsic and extrinsic parameters of the VI sensor 
were plugged in while other parameters used default values without special tuning.

For one session from location B6\_B1 in Table~\ref{seqtable}, the odometry results by the three methods were visualized in Fig.~\ref{viobenchmark}. The figure showed that both OKVIS and VINS-Mono performed fair with this data session. 
However, their performance was much worse than with the EuRoC MAV dataset \cite{burri25012016} on which 
\cite{delmerico2018benchmark} reported ATEs for both methods no greater than 0.3 m.
We believe the main reasons were twofold.
Firstly, our data contained a variety of challenging factors, such as dynamic scenes and illumination changes. 
Secondly, the data was recorded by robots with nearly planar motion which often rendered the metric scale unobservable \cite{wu2017vins}.
In contrast, adding encoder measurements in the extended VINS-Mono method substantially improved the pose estimation precision in terms of RPE,
and the scale consistency as seen from the estimated trajectory lengths by these methods (see the top of Fig.~\ref{viobenchmark}).

\subsection{Benchmark mapping algorithms}

To show that our benchmark is suitable for evaluating mapping algorithms, this section gives ATE and RPE metrics for a mapping method realized in the maplab framework \cite{schneider2018maplab}.
Initialized by pose estimates from the extended VINS-Mono, 
the method administers loop closure (LC) and global bundle adjustment (BA) steps to refine pose and landmark estimates.
Four data sessions captured on expansive floors with walking people and less textured walls were chosen for this test.

The resulting optimized poses were evaluated with the ATE and RPE metrics as tabulated in Table~\ref{tab:mapping}.
From the table, we see that for locations with traveled distances less than 1000 m,
the encoder-aided VIO handled very well.
For greater traveled distances, the extended VINS-Mono was often insufficient in precision.
For all locations, the steps of LC and BA significantly improved the global consistency and local precision of estimated poses.
But these steps were not very helpful for the third session 
which had false associations among the few loops detected by the LC module.
Close inspection showed that some loops were added due to similar appearances of different places.

\begin{table}[]
	\captionsetup{justification=justified}
	\caption{Values of pose metrics for several sessions with the mapping algorithms. Each dataset was named by the local time at its start. evio - encoder aided VIO, opt. - evio + loop closure + bundle adjustment, L - traveled distance, $\Delta$ = 2 sec. And ref. L was computed based off the lidar data, est. L based off the algorithm result. Note opt. for the third session diverged.}
	\setlength\extrarowheight{4pt}{ %
		\begin{tabular}{@{}l@{\hskip 1mm}|l@{\hskip 1mm}|l@{\hskip 1mm}|l@{\hskip 1mm}|l@{\hskip 1mm}|l@{\hskip 1mm}c@{}} %
			\hline
			\textbf{\begin{tabular}[c]{@{}l@{}}Mapping\\  data\end{tabular}}              & \textbf{Algo.} & \textbf{ATE{[}m{]}} & \textbf{\begin{tabular}[c]{@{}l@{}}RTE($\Delta$)\\  {[}m{]}\end{tabular}} & \textbf{\begin{tabular}[c]{@{}l@{}}RRE($\Delta$)\\  {[}$^{\circ}${]}\end{tabular}} & \textbf{\begin{tabular}[c]{@{}l@{}}ref. L{[}m{]}:\\ est. L{[}m{]}\end{tabular}} \\
			\hline
			\multirow{2}{*}{\begin{tabular}[c]{@{}l@{}}2018-09-28\\ \_15-54-50\end{tabular}} & evio           & 1.223               & 0.081               & 0.773                        & 419.62: 415.06                                                                   \\
			\cline{2-6}
			& opt.           & 0.395               & 0.080               & 0.528                        & 413.75: 418.32                                                                   \\
			\hline
			\multirow{2}{*}{\begin{tabular}[c]{@{}l@{}}2018-08-02\\ \_18-42-03\end{tabular}} & evio           & 0.779               & 0.056               & 1.934                        & 711.10: 706.87                                                                   \\
			\cline{2-6}
			& opt.           & 0.438               & 0.047               & 0.473                        & 709.90: 698.29                                                                   \\
			\hline
			\multirow{2}{*}{\begin{tabular}[c]{@{}l@{}}2018-09-12\\ \_15-38-08\end{tabular}} & evio           & 44.890              & 0.085               & 4.380                        & 1957.57: 2008.44                                                                 \\
			\cline{2-6}
			& opt.           & 15.260              & 0.510               & 2.419                        & 2021.57: 1952.83                                                                 \\
			\hline
			\multirow{2}{*}{\begin{tabular}[c]{@{}l@{}}2018-09-12\\ \_14-43-45\end{tabular}} & evio           & 23.651              & 3.658               & 1.129                        & 1047.35: 1047.19                                                                 \\
			\cline{2-6}
			& opt.           & 2.885               & 0.082               & 0.604                        & 1056.30: 1024.77\\
			\hline                                                       
	\end{tabular}}
	\label{tab:mapping}
\end{table}

\subsection{Benchmark metric place recognition algorithms}

Using the proposed metrics, this section showcases the evaluation of a metric 
place recognition method in the maplab framework \cite{schneider2018maplab} on our dataset.

For two locations in Table ~\ref{seqtable}, we chose two mapping sessions to create summary maps.
A mapping session was selected such that its course of data collection mostly covered those of the other sessions of the location.
For each location, multiple test sessions spanning half a year were selected.
The place recognition was attempted for frames sampled at 1 Hz from every test session.
If a frame was localized by the place recognition module and its estimated pose passed a geometric check \cite{lynen2017trajectory}, it was counted for a localization.

The localization results with the proposed metrics were tabulated in Table~\ref{tab:loc_stat}, showing several interesting facts.
As the time gap between mapping and test sessions grew, 
the normalized localizations did not show an obvious decreasing tendency, 
implying that the time gap did not impact much recall of place recognition.
In contrast, the time of day had a major impact on place recognition 
for data captured in the ground floor of B2 (B2\_F1 in Table~\ref{seqtable}).
The dusk at 17:30 dimming the glass doors and ceilings literally eliminated localizations.
Unsurprisingly, this effect of the time of day was unobvious for sessions at the basement of B6 which usually had lights on all day long.

\begin{figure}[]
	\centering
    \subcaptionbox{}{\includegraphics[width=0.45\columnwidth,trim={0 2mm 0 5.3mm},clip]{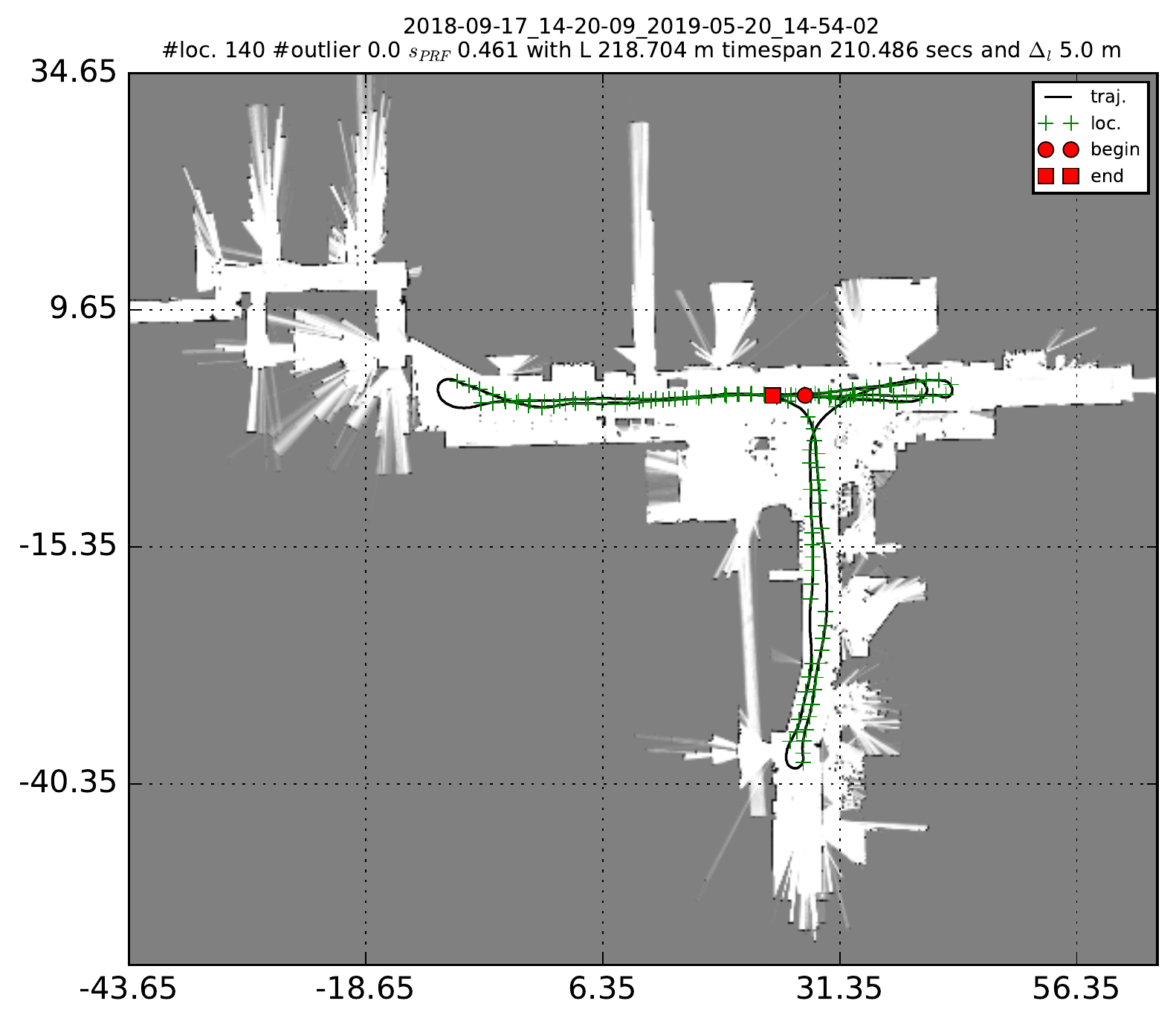}}
	\hfil
	\subcaptionbox{}{\includegraphics[width=0.50\columnwidth,trim={0 8mm 0 0},clip]{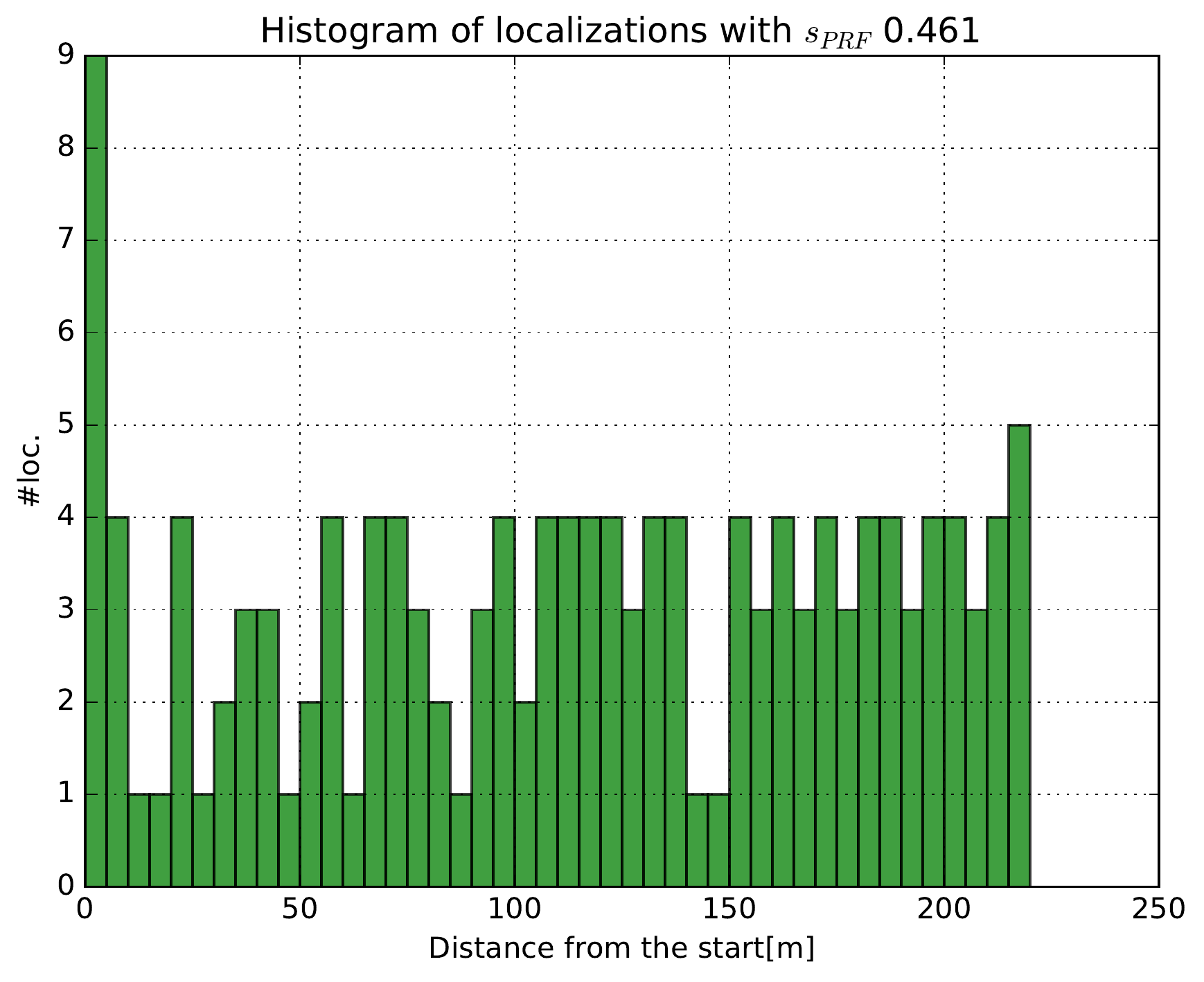}}

    \subcaptionbox{}{\includegraphics[width=0.45\columnwidth,trim={0 2mm 0 5.3mm},clip]{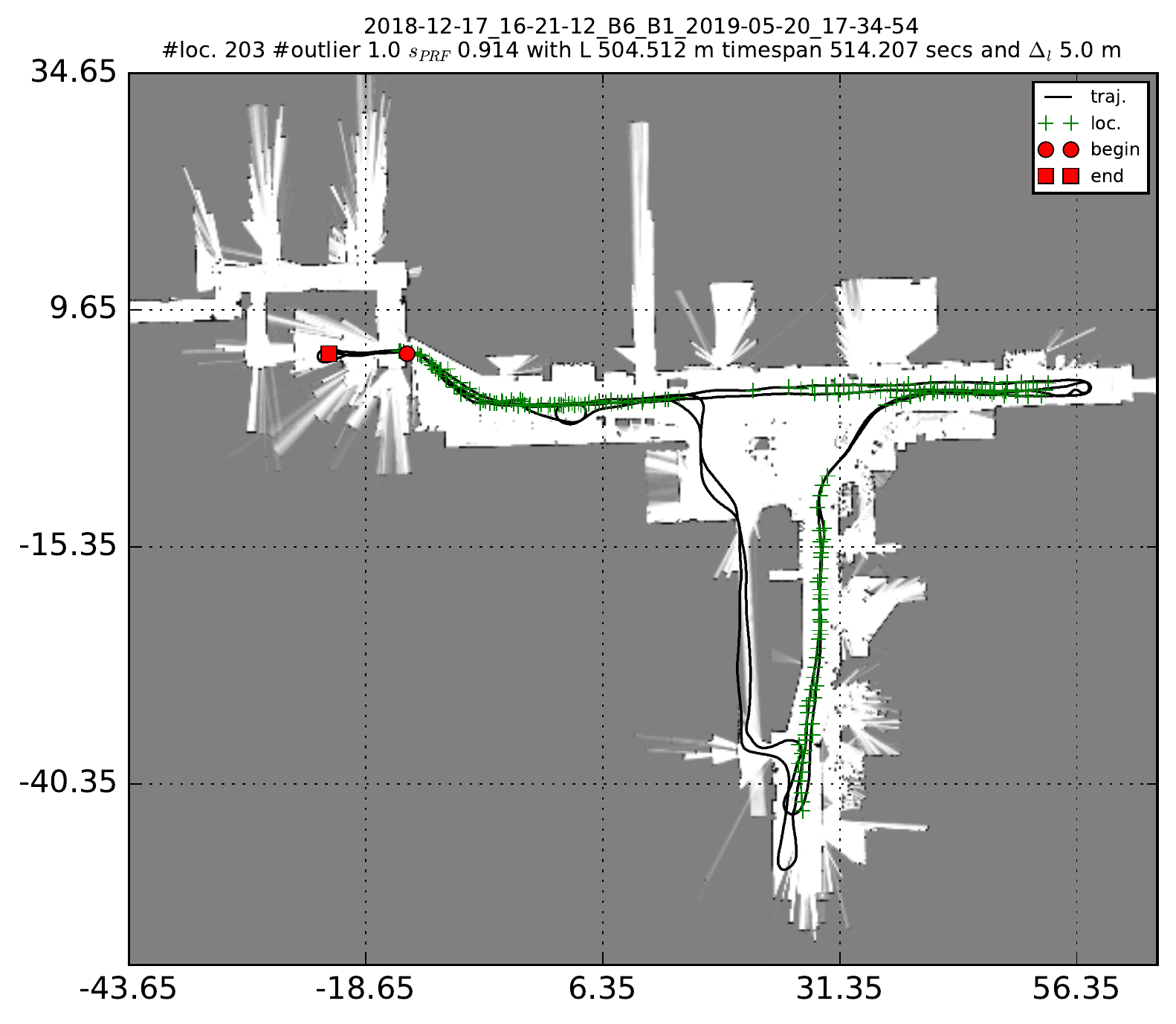}}
    \hfil
	\subcaptionbox{}{\includegraphics[width=0.50\columnwidth,trim={0 8mm 0 0},clip]{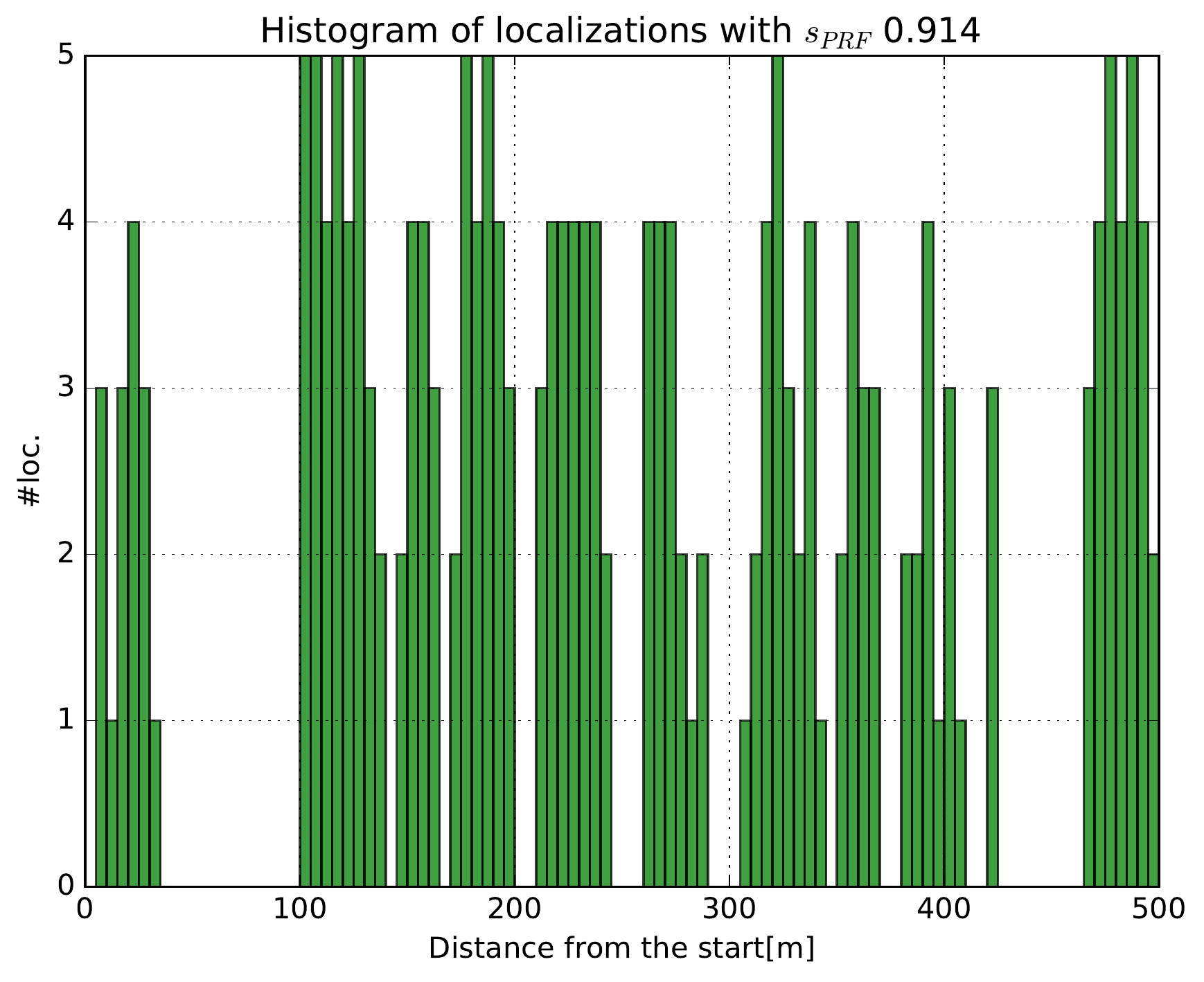}}

    \subcaptionbox{}{\includegraphics[width=0.45\columnwidth,trim={0 2mm 0 5.3mm},clip]{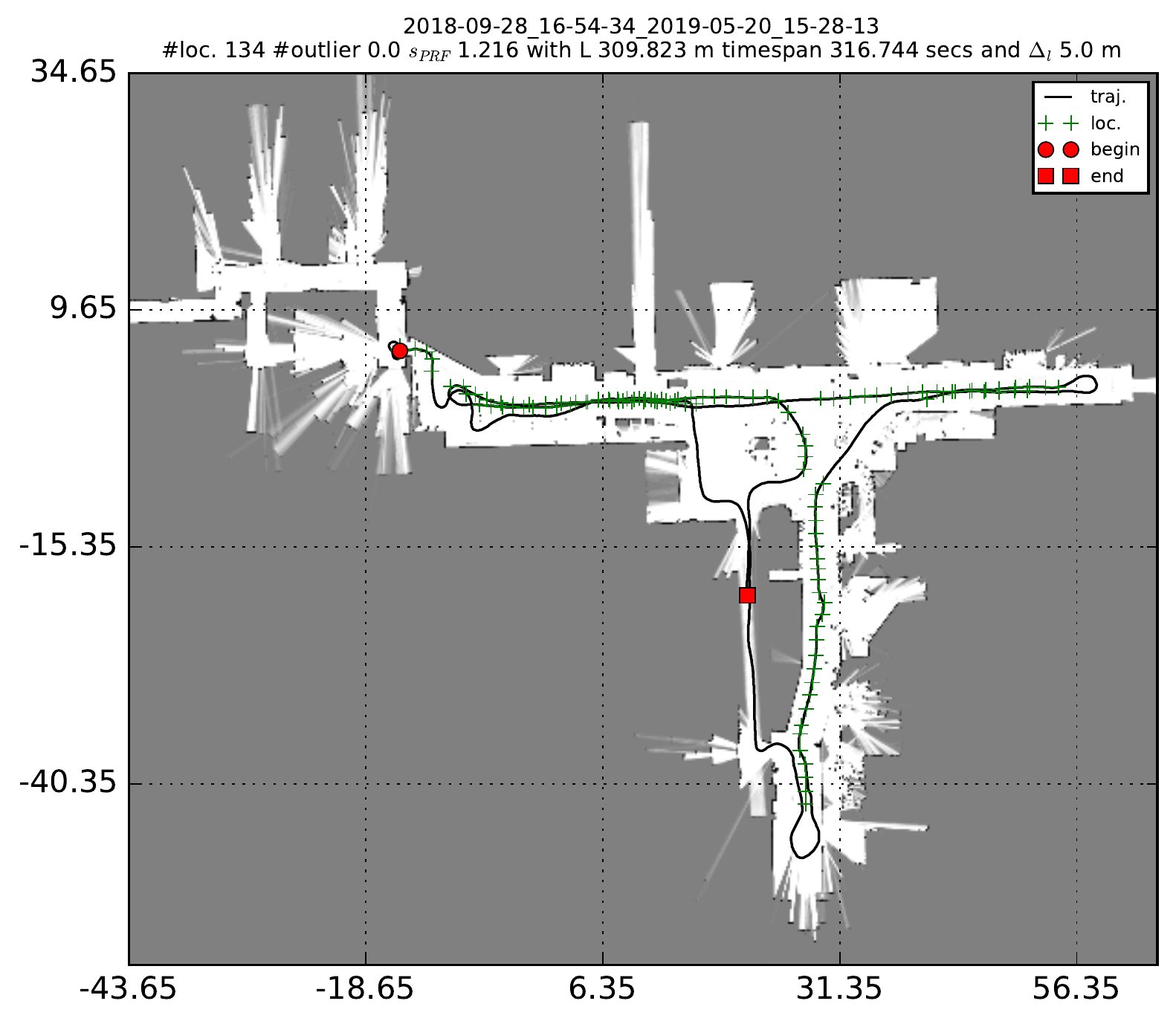}}
    \hfil
	\subcaptionbox{}{\includegraphics[width=0.50\columnwidth,trim={0 8mm 0 0},clip]{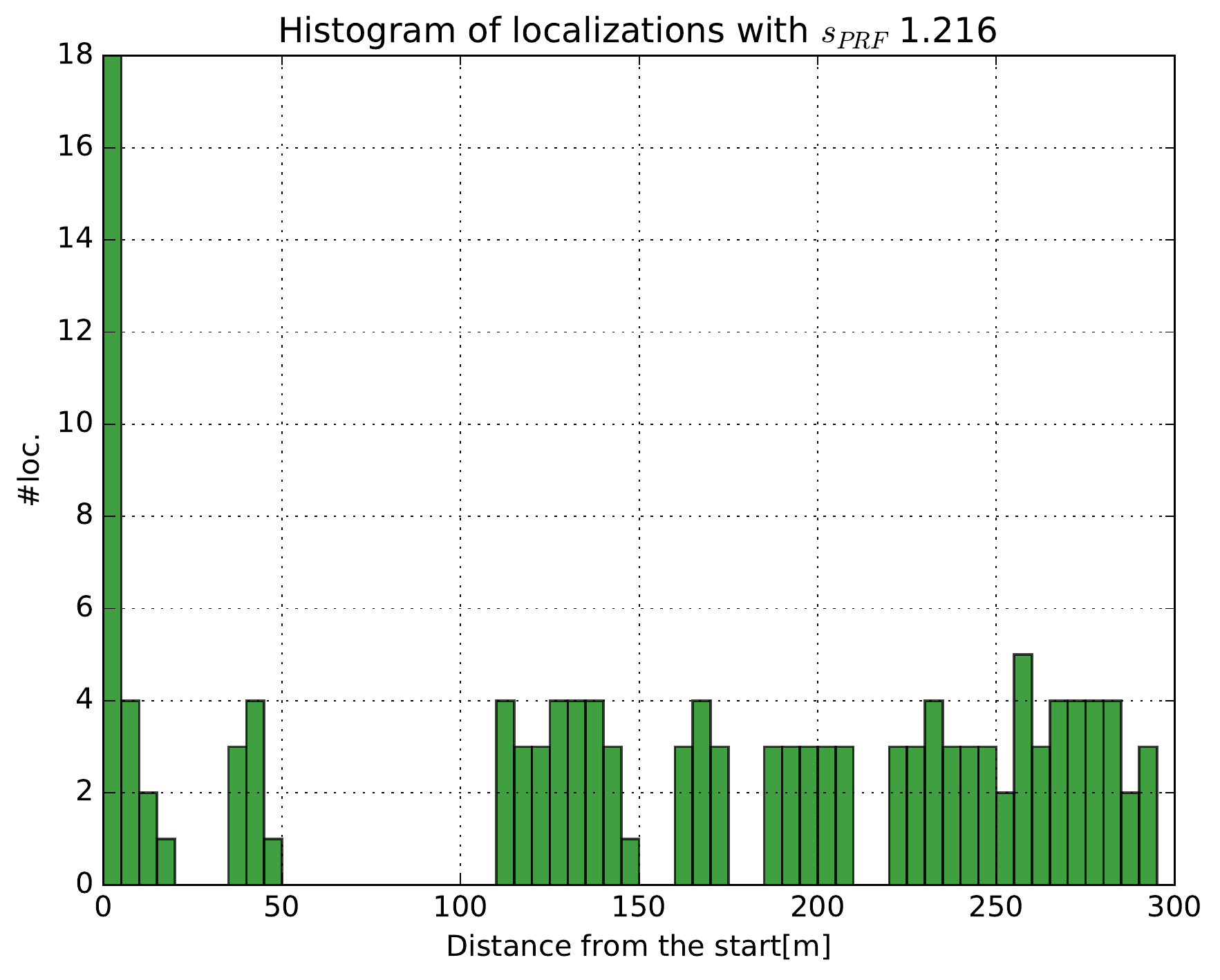}}

    \caption{The $s_{PRF}$ indicates the regularity of localizations over distance.
        Each row shows the localizations in the pre-built map on the left and
        the histogram over traveled distance on the right for one test session.
        Except for the vertical axis of the histogram, all axes have a unit of meter.
    }
	\label{fig:s_prf}
\end{figure}

Moreover, Fig.~\ref{fig:s_prf} illustrated how $s_{PRF}$ describes the distribution of localizations. From top to bottom of the Figure,
 as $s_{PRF}$ grew from 0.416 to 1.216, the occurrences became less regular, 
 showing that greater values of $s_{PRF}$ were associated with less regular occurrences.
The fact that $s_{PRF}$ captured the localization regularity was further confirmed by the complete list of figures for tests in Table~\ref{tab:loc_stat} provided on the benchmark website.

\begin{table*}[!t]
	\renewcommand{\arraystretch}{1.5}
	\caption{Localization statistics for two sample data groups. $s_{PRF}$ - standard deviation of the place recognition frequencies, $L$ - traveled distance, $t_n$ - dataset duration, $N_{tp}$ - number of true positive localizations = \#loc. - \#outlier. Note each data session is named by the local time of its start.}
	\centering
	\begin{tabular}{cllllllll}
		\hline
		\multicolumn{1}{c|}{\bfseries Mapping dataset}                                                                                              & \multicolumn{1}{c|}{\bfseries Localization dataset} & \multicolumn{1}{c|}{\bfseries \#loc.} & \multicolumn{1}{c|}{\bfseries \#outlier} & \multicolumn{1}{c|}{\boldmath{$s_{PRF}$}} & \multicolumn{1}{c|}{\bfseries \boldmath{$L$}{[}m{]}} & \multicolumn{1}{c|}{\boldmath{$t_n$}\bfseries {[}sec{]}} & \multicolumn{1}{c|}{\boldmath{$N_{tp}/L$}} & \multicolumn{1}{c}{\boldmath{$N_{tp}/t_n$}} \\ \hline\hline
		\multicolumn{1}{c|}{\multirow{17}{*}{\begin{tabular}[c]{@{}c@{}}Ground floor B2 building:\\ 2018-09-28\_15-54-50\end{tabular}}}   & \multicolumn{1}{l|}{2018-09-21\_14-46-46} & \multicolumn{1}{l|}{605}    & \multicolumn{1}{l|}{2.5}       & \multicolumn{1}{l|}{1.313}     & \multicolumn{1}{l|}{500.90}   & \multicolumn{1}{l|}{1077.07}        & \multicolumn{1}{l|}{1.203}              & 0.559                                    \\ \cline{2-9} 
		\multicolumn{1}{c|}{}                                                                                                             & \multicolumn{1}{l|}{2018-09-28\_15-54-50} & \multicolumn{1}{l|}{572}    & \multicolumn{1}{l|}{1}         & \multicolumn{1}{l|}{0.420}     & \multicolumn{1}{l|}{419.53}   & \multicolumn{1}{l|}{712.67}         & \multicolumn{1}{l|}{1.361}              & 0.801                                    \\ \cline{2-9} 
		\multicolumn{1}{c|}{}                                                                                                             & \multicolumn{1}{l|}{2018-10-12\_07-05-30} & \multicolumn{1}{l|}{434}    & \multicolumn{1}{l|}{1}         & \multicolumn{1}{l|}{0.299}     & \multicolumn{1}{l|}{393.67}   & \multicolumn{1}{l|}{591.55}         & \multicolumn{1}{l|}{1.100}              & 0.732                                    \\ \cline{2-9} 
		\multicolumn{1}{c|}{}                                                                                                             & \multicolumn{1}{l|}{2018-11-01\_15-36-43} & \multicolumn{1}{l|}{396}    & \multicolumn{1}{l|}{1}         & \multicolumn{1}{l|}{0.348}     & \multicolumn{1}{l|}{393.94}   & \multicolumn{1}{l|}{598.08}         & \multicolumn{1}{l|}{1.003}              & 0.660                                    \\ \cline{2-9} 
		\multicolumn{1}{c|}{}                                                                                                             & \multicolumn{1}{l|}{2018-11-15\_15-06-32} & \multicolumn{1}{l|}{325}    & \multicolumn{1}{l|}{2}         & \multicolumn{1}{l|}{0.681}     & \multicolumn{1}{l|}{403.02}   & \multicolumn{1}{l|}{693.22}         & \multicolumn{1}{l|}{0.801}              & 0.466                                    \\ \cline{2-9} 
		\multicolumn{1}{c|}{}                                                                                                             & \multicolumn{1}{l|}{2018-11-19\_15-12-44} & \multicolumn{1}{l|}{111}    & \multicolumn{1}{l|}{4.5}       & \multicolumn{1}{l|}{1.114}     & \multicolumn{1}{l|}{334.55}   & \multicolumn{1}{l|}{377.55}         & \multicolumn{1}{l|}{0.318}              & 0.282                                    \\ \cline{2-9} 
		\multicolumn{1}{c|}{}                                                                                                             & \multicolumn{1}{l|}{2018-11-19\_15-35-14} & \multicolumn{1}{l|}{53}     & \multicolumn{1}{l|}{1}         & \multicolumn{1}{l|}{1.430}     & \multicolumn{1}{l|}{242.24}   & \multicolumn{1}{l|}{254.71}         & \multicolumn{1}{l|}{0.215}              & 0.204                                    \\ \cline{2-9} 
		\multicolumn{1}{c|}{}                                                                                                             & \multicolumn{1}{l|}{2018-11-19\_15-40-58} & \multicolumn{1}{l|}{137}    & \multicolumn{1}{l|}{1}         & \multicolumn{1}{l|}{1.040}     & \multicolumn{1}{l|}{340.45}   & \multicolumn{1}{l|}{382.51}         & \multicolumn{1}{l|}{0.399}              & 0.356                                    \\ \cline{2-9} 
		\multicolumn{1}{c|}{}                                                                                                             & \multicolumn{1}{l|}{2018-11-20\_14-58-12} & \multicolumn{1}{l|}{179}    & \multicolumn{1}{l|}{1}         & \multicolumn{1}{l|}{1.044}     & \multicolumn{1}{l|}{488.74}   & \multicolumn{1}{l|}{510.01}         & \multicolumn{1}{l|}{0.364}              & 0.349                                    \\ \cline{2-9} 
		\multicolumn{1}{c|}{}                                                                                                             & \multicolumn{1}{l|}{2018-11-22\_14-15-54} & \multicolumn{1}{l|}{280}    & \multicolumn{1}{l|}{2.5}       & \multicolumn{1}{l|}{0.934}     & \multicolumn{1}{l|}{817.01}   & \multicolumn{1}{l|}{700.09}         & \multicolumn{1}{l|}{0.340}              & 0.396                                    \\ \cline{2-9} 
		\multicolumn{1}{c|}{}                                                                                                             & \multicolumn{1}{l|}{2018-11-22\_14-30-03} & \multicolumn{1}{l|}{35}     & \multicolumn{1}{l|}{1.5}       & \multicolumn{1}{l|}{1.263}     & \multicolumn{1}{l|}{147.78}   & \multicolumn{1}{l|}{131.86}         & \multicolumn{1}{l|}{0.227}              & 0.254                                    \\ \cline{2-9} 
		\multicolumn{1}{c|}{}                                                                                                             & \multicolumn{1}{l|}{2018-11-22\_17-29-21} & \multicolumn{1}{l|}{2}      & \multicolumn{1}{l|}{0}         & \multicolumn{1}{l|}{4.743}     & \multicolumn{1}{l|}{232.11}   & \multicolumn{1}{l|}{208.06}         & \multicolumn{1}{l|}{0.009}              & 0.010                                    \\ \cline{2-9} 
		\multicolumn{1}{c|}{}                                                                                                             & \multicolumn{1}{l|}{2018-11-22\_17-33-32} & \multicolumn{1}{l|}{0}      & \multicolumn{1}{l|}{0}         & \multicolumn{1}{l|}{0.000}     & \multicolumn{1}{l|}{221.07}   & \multicolumn{1}{l|}{212.18}         & \multicolumn{1}{l|}{0.000}              & 0.000                                    \\ \cline{2-9} 
		\multicolumn{1}{c|}{}                                                                                                             & \multicolumn{1}{l|}{2018-11-29\_15-18-53} & \multicolumn{1}{l|}{213}    & \multicolumn{1}{l|}{1}         & \multicolumn{1}{l|}{0.781}     & \multicolumn{1}{l|}{407.20}   & \multicolumn{1}{l|}{455.29}         & \multicolumn{1}{l|}{0.521}              & 0.466                                    \\ \cline{2-9} 
		\multicolumn{1}{c|}{}                                                                                                             & \multicolumn{1}{l|}{2018-12-05\_14-25-56} & \multicolumn{1}{l|}{330}    & \multicolumn{1}{l|}{4.5}       & \multicolumn{1}{l|}{0.571}     & \multicolumn{1}{l|}{555.06}   & \multicolumn{1}{l|}{562.79}         & \multicolumn{1}{l|}{0.586}              & 0.578                                    \\ \cline{2-9} 
		\multicolumn{1}{c|}{}                                                                                                             & \multicolumn{1}{l|}{2018-12-21\_15-28-22} & \multicolumn{1}{l|}{473}    & \multicolumn{1}{l|}{1.5}       & \multicolumn{1}{l|}{0.673}     & \multicolumn{1}{l|}{594.50}   & \multicolumn{1}{l|}{740.71}         & \multicolumn{1}{l|}{0.793}              & 0.637                                    \\ \cline{2-9} 
		\multicolumn{1}{c|}{}                                                                                                             & \multicolumn{1}{l|}{2018-12-27\_15-46-23} & \multicolumn{1}{l|}{135}    & \multicolumn{1}{l|}{6.5}       & \multicolumn{1}{l|}{1.069}     & \multicolumn{1}{l|}{438.57}   & \multicolumn{1}{l|}{428.32}         & \multicolumn{1}{l|}{0.293}              & 0.300                                    \\ \hline\hline
		
		\multicolumn{1}{c|}{\multirow{26}{*}{\begin{tabular}[c]{@{}c@{}}First basement B6 building:\\ 2018-08-02\_18-42-03\end{tabular}}} & \multicolumn{1}{l|}{2018-08-02\_18-24-32} & \multicolumn{1}{l|}{221}    & \multicolumn{1}{l|}{0}         & \multicolumn{1}{l|}{0.254}     & \multicolumn{1}{l|}{266.60}   & \multicolumn{1}{l|}{308.02}         & \multicolumn{1}{l|}{0.829}              & 0.717                                    \\ \cline{2-9} 
		\multicolumn{1}{c|}{}                                                                                                             & \multicolumn{1}{l|}{2018-08-02\_18-42-03} & \multicolumn{1}{l|}{624}    & \multicolumn{1}{l|}{0}         & \multicolumn{1}{l|}{0.371}     & \multicolumn{1}{l|}{700.88}   & \multicolumn{1}{l|}{752.53}         & \multicolumn{1}{l|}{0.890}              & 0.829                                    \\ \cline{2-9} 
		\multicolumn{1}{c|}{}                                                                                                             & \multicolumn{1}{l|}{2018-09-07\_14-22-04} & \multicolumn{1}{l|}{3738}   & \multicolumn{1}{l|}{3}         & \multicolumn{1}{l|}{0.452}     & \multicolumn{1}{l|}{3671.08}  & \multicolumn{1}{l|}{7412.40}        & \multicolumn{1}{l|}{1.017}              & 0.504                                    \\ \cline{2-9} 
		\multicolumn{1}{c|}{}                                                                                                             & \multicolumn{1}{l|}{2018-09-17\_14-04-30} & \multicolumn{1}{l|}{128}    & \multicolumn{1}{l|}{0}         & \multicolumn{1}{l|}{0.538}     & \multicolumn{1}{l|}{217.44}   & \multicolumn{1}{l|}{263.80}         & \multicolumn{1}{l|}{0.589}              & 0.485                                    \\ \cline{2-9} 
		\multicolumn{1}{c|}{}                                                                                                             & \multicolumn{1}{l|}{2018-09-17\_14-10-00} & \multicolumn{1}{l|}{138}    & \multicolumn{1}{l|}{0}         & \multicolumn{1}{l|}{0.492}     & \multicolumn{1}{l|}{212.35}   & \multicolumn{1}{l|}{231.86}         & \multicolumn{1}{l|}{0.650}              & 0.595                                    \\ \cline{2-9} 
		\multicolumn{1}{c|}{}                                                                                                             & \multicolumn{1}{l|}{2018-09-17\_14-14-58} & \multicolumn{1}{l|}{46}     & \multicolumn{1}{l|}{1.5}       & \multicolumn{1}{l|}{0.745}     & \multicolumn{1}{l|}{86.41}    & \multicolumn{1}{l|}{94.67}          & \multicolumn{1}{l|}{0.515}              & 0.470                                    \\ \cline{2-9} 
		\multicolumn{1}{c|}{}                                                                                                             & \multicolumn{1}{l|}{2018-09-17\_14-20-09} & \multicolumn{1}{l|}{140}    & \multicolumn{1}{l|}{0}         & \multicolumn{1}{l|}{0.461}     & \multicolumn{1}{l|}{218.70}   & \multicolumn{1}{l|}{210.49}         & \multicolumn{1}{l|}{0.640}              & 0.665                                    \\ \cline{2-9} 
		\multicolumn{1}{c|}{}                                                                                                             & \multicolumn{1}{l|}{2018-09-17\_15-22-19} & \multicolumn{1}{l|}{974}    & \multicolumn{1}{l|}{0}         & \multicolumn{1}{l|}{0.510}     & \multicolumn{1}{l|}{938.85}   & \multicolumn{1}{l|}{1470.47}        & \multicolumn{1}{l|}{1.037}              & 0.662                                    \\ \cline{2-9} 
		\multicolumn{1}{c|}{}                                                                                                             & \multicolumn{1}{l|}{2018-09-19\_14-30-53} & \multicolumn{1}{l|}{187}    & \multicolumn{1}{l|}{0}         & \multicolumn{1}{l|}{0.460}     & \multicolumn{1}{l|}{187.97}   & \multicolumn{1}{l|}{311.35}         & \multicolumn{1}{l|}{0.995}              & 0.601                                    \\ \cline{2-9} 
		\multicolumn{1}{c|}{}                                                                                                             & \multicolumn{1}{l|}{2018-09-28\_16-54-34} & \multicolumn{1}{l|}{134}    & \multicolumn{1}{l|}{0}         & \multicolumn{1}{l|}{1.216}     & \multicolumn{1}{l|}{309.82}   & \multicolumn{1}{l|}{316.74}         & \multicolumn{1}{l|}{0.433}              & 0.423                                    \\ \cline{2-9} 
		\multicolumn{1}{c|}{}                                                                                                             & \multicolumn{1}{l|}{2018-09-28\_17-05-32} & \multicolumn{1}{l|}{118}    & \multicolumn{1}{l|}{0}         & \multicolumn{1}{l|}{0.659}     & \multicolumn{1}{l|}{222.77}   & \multicolumn{1}{l|}{242.95}         & \multicolumn{1}{l|}{0.530}              & 0.486                                    \\ \cline{2-9} 
		\multicolumn{1}{c|}{}                                                                                                             & \multicolumn{1}{l|}{2018-09-30\_15-20-22} & \multicolumn{1}{l|}{420}    & \multicolumn{1}{l|}{1}         & \multicolumn{1}{l|}{0.463}     & \multicolumn{1}{l|}{371.52}   & \multicolumn{1}{l|}{612.08}         & \multicolumn{1}{l|}{1.128}              & 0.685                                    \\ \cline{2-9} 
		\multicolumn{1}{c|}{}                                                                                                             & \multicolumn{1}{l|}{2018-10-08\_14-37-26} & \multicolumn{1}{l|}{384}    & \multicolumn{1}{l|}{0}         & \multicolumn{1}{l|}{0.437}     & \multicolumn{1}{l|}{373.45}   & \multicolumn{1}{l|}{643.71}         & \multicolumn{1}{l|}{1.028}              & 0.597                                    \\ \cline{2-9} 
		\multicolumn{1}{c|}{}                                                                                                             & \multicolumn{1}{l|}{2018-10-17\_06-49-18} & \multicolumn{1}{l|}{321}    & \multicolumn{1}{l|}{1.5}       & \multicolumn{1}{l|}{0.646}     & \multicolumn{1}{l|}{378.22}   & \multicolumn{1}{l|}{692.05}         & \multicolumn{1}{l|}{0.845}              & 0.462                                    \\ \cline{2-9} 
		\multicolumn{1}{c|}{}                                                                                                             & \multicolumn{1}{l|}{2018-10-25\_15-52-55} & \multicolumn{1}{l|}{357}    & \multicolumn{1}{l|}{2}         & \multicolumn{1}{l|}{1.333}     & \multicolumn{1}{l|}{367.72}   & \multicolumn{1}{l|}{645.10}         & \multicolumn{1}{l|}{0.965}              & 0.550                                    \\ \cline{2-9} 
		\multicolumn{1}{c|}{}                                                                                                             & \multicolumn{1}{l|}{2018-11-12\_16-49-53} & \multicolumn{1}{l|}{329}    & \multicolumn{1}{l|}{0}         & \multicolumn{1}{l|}{0.533}     & \multicolumn{1}{l|}{382.13}   & \multicolumn{1}{l|}{575.08}         & \multicolumn{1}{l|}{0.861}              & 0.572                                    \\ \cline{2-9} 
		\multicolumn{1}{c|}{}                                                                                                             & \multicolumn{1}{l|}{2018-11-12\_16-59-51} & \multicolumn{1}{l|}{319}    & \multicolumn{1}{l|}{1}         & \multicolumn{1}{l|}{0.536}     & \multicolumn{1}{l|}{381.61}   & \multicolumn{1}{l|}{552.93}         & \multicolumn{1}{l|}{0.833}              & 0.575                                    \\ \cline{2-9} 
		\multicolumn{1}{c|}{}                                                                                                             & \multicolumn{1}{l|}{2018-11-12\_17-31-23} & \multicolumn{1}{l|}{42}     & \multicolumn{1}{l|}{0}         & \multicolumn{1}{l|}{0.924}     & \multicolumn{1}{l|}{70.27}    & \multicolumn{1}{l|}{161.07}         & \multicolumn{1}{l|}{0.598}              & 0.261                                    \\ \cline{2-9} 
		\multicolumn{1}{c|}{}                                                                                                             & \multicolumn{1}{l|}{2018-11-12\_17-49-49} & \multicolumn{1}{l|}{38}     & \multicolumn{1}{l|}{0}         & \multicolumn{1}{l|}{0.829}     & \multicolumn{1}{l|}{68.66}    & \multicolumn{1}{l|}{117.61}         & \multicolumn{1}{l|}{0.553}              & 0.323                                    \\ \cline{2-9} 
		\multicolumn{1}{c|}{}                                                                                                             & \multicolumn{1}{l|}{2018-11-12\_17-53-43} & \multicolumn{1}{l|}{148}    & \multicolumn{1}{l|}{1}         & \multicolumn{1}{l|}{0.555}     & \multicolumn{1}{l|}{192.78}   & \multicolumn{1}{l|}{283.98}         & \multicolumn{1}{l|}{0.763}              & 0.518                                    \\ \cline{2-9} 
		\multicolumn{1}{c|}{}                                                                                                             & \multicolumn{1}{l|}{2018-11-19\_14-41-17} & \multicolumn{1}{l|}{316}    & \multicolumn{1}{l|}{0}         & \multicolumn{1}{l|}{0.537}     & \multicolumn{1}{l|}{389.49}   & \multicolumn{1}{l|}{636.08}         & \multicolumn{1}{l|}{0.811}              & 0.497                                    \\ \cline{2-9} 
		\multicolumn{1}{c|}{}                                                                                                             & \multicolumn{1}{l|}{2018-11-28\_14-57-09} & \multicolumn{1}{l|}{624}    & \multicolumn{1}{l|}{0}         & \multicolumn{1}{l|}{0.571}     & \multicolumn{1}{l|}{729.83}   & \multicolumn{1}{l|}{1124.15}        & \multicolumn{1}{l|}{0.855}              & 0.555                                    \\ \cline{2-9} 
		\multicolumn{1}{c|}{}                                                                                                             & \multicolumn{1}{l|}{2018-12-05\_10-51-59} & \multicolumn{1}{l|}{286}    & \multicolumn{1}{l|}{1}         & \multicolumn{1}{l|}{0.717}     & \multicolumn{1}{l|}{566.95}   & \multicolumn{1}{l|}{640.88}         & \multicolumn{1}{l|}{0.503}              & 0.445                                    \\ \cline{2-9} 
		\multicolumn{1}{c|}{}                                                                                                             & \multicolumn{1}{l|}{2018-12-17\_16-16-04} & \multicolumn{1}{l|}{115}    & \multicolumn{1}{l|}{1}         & \multicolumn{1}{l|}{0.669}     & \multicolumn{1}{l|}{201.69}   & \multicolumn{1}{l|}{228.10}         & \multicolumn{1}{l|}{0.565}              & 0.500                                    \\ \cline{2-9} 
		\multicolumn{1}{c|}{}                                                                                                             & \multicolumn{1}{l|}{2018-12-17\_16-21-12} & \multicolumn{1}{l|}{203}    & \multicolumn{1}{l|}{1}         & \multicolumn{1}{l|}{0.914}     & \multicolumn{1}{l|}{504.51}   & \multicolumn{1}{l|}{514.21}         & \multicolumn{1}{l|}{0.400}              & 0.393                                    \\ \cline{2-9} 
		\multicolumn{1}{c|}{}                                                                                                             & \multicolumn{1}{l|}{2018-12-21\_17-13-26} & \multicolumn{1}{l|}{672}    & \multicolumn{1}{l|}{1}         & \multicolumn{1}{l|}{0.812}     & \multicolumn{1}{l|}{859.00}   & \multicolumn{1}{l|}{1299.87}        & \multicolumn{1}{l|}{0.781}              & 0.516                                    \\ \hline
	\end{tabular}
    \label{tab:loc_stat}
\end{table*}

\section{CONCLUSION}
We present a realistic and demanding benchmark for evaluating place recognition and SLAM methods.
Recorded by a fleet of Segway delivery robots each of which outfitted with
a low-cost RealSense VI sensor, two wheel encoders, and a removable Hokuyo 2D lidar for creating reference poses, 
the dataset featured many challenging factors such as dynamic obstacles, illumination change, bumpy floors, and quick turns.
Spanning half a year, the robots recorded data sessions repeatedly traversing the same routes in several indoor locations.
These attributes of the dataset make it suitable for evaluating SLAM and place recognition methods.

We also propose several metrics to evaluate metric place recognition methods
 accounting for the frequency and regularity of localization occurrences.
These metrics and a few metrics for SLAM methods were used to evaluate several state-of-the-art SLAM and place recognition methods on our benchmark, 
showing the fitness of the proposed metrics and the challenges in our dataset.

We are making the evaluation tools an online service as the KITTI benchmark \cite{geiger2012cvpr}.
Over time, more sessions of data for diverse scenes and sensors will be added as the robot delivery service expands.

\bibliographystyle{IEEEtran}
\bibliography{ms}

\end{document}